\documentclass[11pt,a4paper]{article}
\usepackage[hyperref]{emnlp2020}
\usepackage{times}
\usepackage{float}
\usepackage{latexsym}

\usepackage{tikz}
\usepackage{csquotes}
\usepackage{pgfplots}
\usepackage{amsmath}

\usepgfplotslibrary{groupplots}
\usetikzlibrary{pgfplots.groupplots}
\pgfplotsset{compat=1.5}
\usepackage{pgfplots,pgfplotstable}
\usetikzlibrary{patterns}

\usepackage{microtype}
\usepackage{amsmath}
\usepackage{amssymb}
\usepackage{mathtools, nccmath}
\usepackage{graphicx}
\graphicspath{ {./images/} }
\usepackage{caption}
\usepackage{subcaption}
\usepackage{xargs} 
\usepackage{xcolor,soul}
\usepackage{booktabs}
\usepackage{multirow}
\usepackage{arydshln}
\usepackage{centernot}

\definecolor{OliveGreen}{RGB}{128,128,0}
\definecolor{PaleBlue}{RGB}{208,228,245}
\definecolor{Blue}{RGB}{111,168,220}

\newcommand{\fc}{fact-checking}

\newcommandx{\meetingnotes}[2][1=]{\todo[linecolor=blue,backgroundcolor=blue!25,bordercolor=blue,#1]{#2}}
\newcommandx{\info}[2][1=]{\todo[linecolor=cyan,backgroundcolor=cyan!25,bordercolor=cyan,#1]{#2}}

\definecolor{alizarin}{rgb}{0.82, 0.1, 0.26}
\definecolor{darkgreen}{rgb}{0.0, 0.5, 0.0}
\definecolor{amber}{rgb}{1.0, 0.49, 0.0}

\aclfinalcopy 
\def\aclpaperid{3515} 

\title{
Explainable Automated Fact-Checking for Public Health Claims
}
\author{Neema Kotonya \and Francesca Toni \\
  Department of Computing\\
  Imperial College London, United Kingdom\\
  \texttt{\{nk2418,ft\}@ic.ac.uk} \\}
  
\date{}

\begin{document}
\maketitle
\begin{abstract}

Fact-checking is the task of verifying the veracity of claims
by assessing their assertions against credible evidence. The vast majority of fact-checking studies focus exclusively on political claims. Very little research explores 
fact-checking for other topics, specifically subject matters for which expertise is required. We present the first study of explainable fact-checking for claims which require specific expertise. For our case study we choose the
setting of public health. To support this case study we construct a new dataset \textsc{PubHealth} of 11.8K claims 
accompanied by
 journalist crafted, gold standard explanations (i.e., judgments) to support the fact-check labels for claims\footnote{Data and code are available here: \url{https://github.com/neemakot/Health-Fact-Checking}}.
We explore two tasks: veracity prediction and explanation generation.
We also define and evaluate, with humans and computationally, three \emph{coherence} properties of explanation quality. Our results indicate that, by training on in-domain data, gains can be made in explainable, automated fact-checking for claims which require specific expertise.
\end{abstract}

\section{Introduction}
\label{sec:introduction}

A great amount of progress has been made in the area of automated fact-checking. 
This includes more accurate machine learning models for veracity prediction and datasets of both naturally occurring \cite{wang-2017-liar,augenstein-etal-2019-multifc,hanselowski-etal-2019-richly} and human-crafted  \cite{thorne-etal-2018-fever} fact-checking claims, against which
the models can be evaluated. However, a few blind spots exist in the state-of-the-art. In this work we address specifically two 
shortcomings: the narrow focus on political claims, and the paucity of explainable systems. 

One subject area which we believe could benefit from expertise-based fact-checking is public health -- including the study of epidemiology, disease prevention in a population, and the formulation of public policies \cite{turnock2012public}.
Recent events, including the COVID-19 pandemic, demonstrate the significant potential harm of misinformation in the public health setting, and the importance in accurately fact-checking claims. Unlike political and general misinformation, specific expertise is required in order to fact check claims in this domain. Oftentimes this expertise 
may be limited, and 
thus claims which surround public health  may be inaccessible (e.g., because of the use of jargon and biomedical terminology) in a way political claims are not. 
Nonetheless, like political misinformation, the public health variety is also potentially very dangerous, because it can put people in imminent danger and risk lives.

Typically, statements which are candidates for fact-checking 
originate in the political domain \cite{vlachos-riedel-2014-fact,ferreira-vlachos-2016-emergent,wang-2017-liar}, and  
tend to surround more general topics or be non-subject specific
\cite{thorne-etal-2018-fever}. This follows the trend of the rising interest in political fact-checking in the last decade \cite{graves2018boundaries}. There are on-going efforts with respect to fact-checking scientific claims \cite{grabitz2017science}. Fact-checking in domains where specific subject expertise is required presents an interesting challenge because general purpose fact-checking systems will not necessarily adapt well to these domains.

The second shortcoming we look to address is the paucity of explainable models for fact-checking (of any kind). Explanations have a particularly important role to play in the task of automated fact-checking. The efficacy of journalistic fact-checking hinges on the credibility and reliability of the fact-check, and explanations (e.g., provided by model agnostic tools such as LIME \cite{ribeiro2016should}) can strengthen this by communicating fidelity in predictive models.
Explainable models can also aid the end users' understanding as they further elucidate claims and their context. 

In this study we explore the novel case of explainable automated fact-checking for claims for which specialised expertise or in-domain knowledge is essential. For our case study we examine the the public health (biomedical) context.

The system for veracity prediction we aim to produce must fulfil two requirements: (1) it should provide a human-understandable explanation (i.e., judgment) for the fact-checking prediction, and (2) that judgement should be understandable for  people who do not have expertise in the subject domain.
We list the following as our \textbf{three main contributions} in this paper:

\begin{enumerate}
\item We present a novel dataset for explainable fact-checking with gold standard fact-checking explanations by journalists. To the best of our knowledge, this is the first dataset specifically for fact-checking in the public health setting.
\item We introduce a framework for generating explanations and veracity prediction specific to public health fact-checking. We show that gains can be made through the use of in-domain data.
\item In order to evaluate the quality of our fact-checking explanations, we define three coherence properties. These can be evaluated by humans as well as computationally, as approximations for human evaluations of fact-checking explanations.
\end{enumerate}

The explanation model trained on in-domain data outperforms the general purpose model on 
summarization evaluation and also when evaluated for explanation quality.

\section{Related Work}
\label{sec:related-work}
A number of recent works in automated fact-checking look at various formulations of fact-checking and its analogous tasks \cite{ferreira-vlachos-2016-emergent,hassan2017toward,zlatkova-etal-2019-fact}. In this paper, we choose to focus on the two specific aspects of concern to us, which have not been thoroughly explored in the literature. These are 
domain-specific and expertise-based claim verification and explainability for automated \fc{} predictions.

\subsection{Language Representations for Health} 

Fewer language resources exist for medical and scientific applications of NLP
compared with other NLP application settings, e.g., social media analysis, NLP for law, and computational journalism and fact-checking. We consider the former below.

There are a number of open source pre-trained language models for NLP applications in the scientific and biomedical domains. The most recent of these pre-trained models are based on the BERT language model \cite{devlin-etal-2019-bert}. 
One example is \textsc{BioBERT}, which is fine-tuned for the biomedical setting \cite{lee2020biobert}. \textsc{BioBERT} is trained on abstracts from PubMed and full article texts from PubMed Central. \textsc{BioBERT} demonstrates higher accuracies when compared to BERT for named entity recognition, relation extraction and question answering in the biomedical domain.

\textsc{SciBERT} is another BERT-based pre-trained model \cite{beltagy-etal-2019-scibert}. \textsc{SciBERT} is trained on 1.14M Semantic Scholar articles relating to computer science and biomedical sciences. Similar to \textsc{BioBERT}, \textsc{SciBERT} also shows improvements on original BERT for in-domain tasks. \textsc{SciBERT} outperforms BERT in five NLP tasks including named entity recognition and text classification.

Given that models for applications of NLP tasks in the biomedical domain, e.g., question answering, show marked improvement when domain-specific, we hypothesize that public health fact-checking could also benefit from the language representations suited for that specific domain. We will make use of both \textsc{SciBERT} and \textsc{BioBERT} in our framework.

\subsection{Explainable Fact-Checking.}

A number of in-roads have been made in developing models to extract explanations from  automated fact-checking systems. To our knowledge, the current state of the art in explainable fact-checking mostly looks to produce extractive explanations, i.e., explanations for veracity predictions in relation to inputs to the system.
Instead, our focus in this paper is on abstractive explanations. We choose this approach, which aims to distill the explanation into the most salient components which form it, as more amenable to users with limited domain expertise, as we discuss below.

Various methods have been applied to the explainable fact-checking task. These methods span the gamut form logic-based approaches such as probabilistic answer set programming \cite{Ahmadi:2019} and reasoning with Horn rules \cite{Ahmadi:2019,gad2019exfakt} to deep learning and attention-based approaches, e.g., leveraging co-attention networks and human annotations in the form of news article comments \cite{Shu:2019}. The outputs of these systems also take a number of forms including Horn rules \cite{Ahmadi:2019}, 
saliency maps \cite{Shu:2019, Popat:2018}, and natural language generation \cite{Atanasova:2020}. 

All approaches produce explanations which are a distillation of the most relevant portion of the system input. In this paper we expand on the work by \citeauthor{Atanasova:2020} as we formulate explanation generation as a summarization exercise.  However, our work differs from the 
existing literature as we construct a framework for joint extractive and \emph{abstractive} explanation generation, as opposed to a purely extractive model. We choose an abstractive approach as we hypothesize that particularly in the case of public health claims, where specific expertise is required to understand the context, abstractive explanations can make the explanation more accessible, particularly for those with little knowledge of the subject matter. In this way we take into account the nature of the claims, something other explainable fact-checking systems do not consider.

\subsection{Evaluation of Explanation Quality}

Only a few explainable fact-checking systems employ thorough evaluation in order to assess the quality of explanations produced. In the cases where evaluations are provided, these primarily take the form of human evaluation, e.g., enlisting annotators to score the quality of explanations with respect to some properties \cite{Atanasova:2020,gad2019exfakt} or through the use of an established evaluation metric in the case where explanation generation is modelled as another task \cite{Atanasova:2020}. 

There is also work on the evaluation of explanation quality more broadly, independently of the task for which explanations are sought. Notably, \citeauthor{sokol2019desiderata} (\citeyear{sokol2019desiderata}) present explainability factsheets for evaluating (machine learning) explanations along five axes, including usability. One of the usability criteria discussed by \citeauthor{sokol2019desiderata} is \emph{coherence}, which we use to develop our three explanation quality properties (see Section \ref{sec:explanation-evaluation}). Whereas \citeauthor{sokol2019desiderata} discuss coherence in general, we provide concrete definitions and use them for evaluating our methods for explaining veracity predictions for public health claims.

\section{The \textsc{PubHealth} dataset}
\label{sec:data}

We constructed a dataset of 11,832 claims for \fc{}, which are related a range of health topics including biomedical subjects (e.g., infectious diseases, stem cell research), government healthcare policy (e.g., abortion, mental health, women's health), and other public health-related stories (see \textit{unproven}, \textit{false} and \textit{mixture} examples in Table \ref{tab:example}), along with explanations offered by 
journalists to support veracity labelling of these claims. The claims were collected from two sources: fact-checking websites and news/news review websites. An example dataset entry is shown in Table \ref{tab:example}.

To the best of our knowledge, this is the first fact-checking dataset to explicitly include gold standard texts provided by journalists specifically as explanation of the fact-checking judgment.
We describe below how the data was collected and processed to obtain the final \textsc{PubHealth} dataset, and provide an analysis of the dataset.

\begin{table*}
\centering
\small
\begin{tabular}{p{5cm}cp{8.3cm}}
\toprule
\textbf{Claim} & \textbf{Label} & \textbf{Explanation}\\
\midrule
Blue Buffalo pet food contains unsafe and higher-than-average levels of lead. & \textcolor{blue}{\textbf{UNPROVEN}} & Aside from a single claimant’s lawsuit against Blue Buffalo and an unrelated recall on one variety of Blue Buffalo product in March 2017, we found no credible information suggesting that Blue Buffalo dog food was tested and found to have abnormally high levels of lead.\\
\addlinespace[0.1cm]
\hdashline
\addlinespace[0.1cm]
Children who watch at least 30 minutes of “Peppa Pig” per day have a 56 percent higher probability of developing \textbf{autism}. & \textcolor{alizarin}{\textbf{FALSE}} & Talk of a Harvard study linking the popular British children’s show “Peppa Pig” to \textbf{autism} went viral, but neither the study nor the scientist who allegedly published it {exists}.\\
\addlinespace[0.1cm]
\hdashline
\addlinespace[0.1cm]
Expired boxes of cake and pancake mix are dangerously toxic. & \textcolor{brown}{\textbf{MIXTURE}} & What's true: Pancake and cake mixes that contain {mold} can cause \textbf{life-threatening allergic reactions}.\newline What's false: Pancake and cake mixes that have passed their expiration dates are not inherently dangerous to ordinarily \textbf{healthy} people, and the yeast in packaged baking products does not ``over time develops spores."\\
\addlinespace[0.1cm]
\hdashline
\addlinespace[0.1cm]
Families tell U.S. lawmakers of \textbf{heparin} deaths. & \textcolor{darkgreen}{\textbf{TRUE}} & A man who said he lost his wife and a son to reactions from tainted \textbf{heparin} made with ingredients from China urged U.S. lawmakers on Tuesday to protect \textbf{patients} from other unsafe \textbf{drugs}.\\

\bottomrule
\end{tabular}
\caption{Example of claims and explanations for \textsc{PubHealth} dataset  entries. Vocabulary from the public health glossary which are contained in the claims and explanations are in \textbf{bold}.}
\label{tab:example}
\end{table*}

\subsection{Data collection}

Initially, we scraped 39,301 claims, amounting to: 27,578 fact-checked claims from five fact-checking websites (Snopes\footnote{\scriptsize\url{https://www.snopes.com/}}, Politifact\footnote{\scriptsize\url{https://www.politifact.com/}}, TruthorFiction\footnote{\scriptsize\url{https://www.truthorfiction.com/}}, FactCheck\footnote{\scriptsize\url{https://www.factcheck.org/}}, and FullFact\footnote{\scriptsize\url{https://fullfact.org/}}); 9,023 news headline claims from the health section and health tags of Associated Press\footnote{\scriptsize\url{https://apnews.com/}} and Reuters News\footnote{\scriptsize\url{https://uk.reuters.com/news/health}} websites; and 2,700 claims from the news review site Health News Review (HNR)\footnote{\scriptsize{\url{ https://www.healthnewsreview.org/}}}. 

We scraped data for two text fields which are essential for fact-checking: 1) the full text of the fact-checking or news article 
discussing the veracity of the claim, 
and 2) 
the fact-checking justification or news summary as explanation for the veracity label of the claim. We also collected
the URLs of sources cited by the journalists in the fact-checking and news articles. 
For each  URL, in the case where the referenced sources could be accessed and read, we also scraped the source texts.

All claims make reference to articles published between October 19 1995 and May 14 2020. In addition to the claim, article texts, explanation texts, and the date on which the fact-check  or news article was published, we scraped meta-data related to each claim. These meta-data include the tags (single or multiple tokens) which may, for example, categorize the topics of the claim or indicate the source of the claim (see Appendix \ref{appendix:labels}), and the names of the fact-checkers and news reporters who contributed to the article.

\subsection{Data processing and analysis}

The data processing involved three tasks: standardizing the veracity labels, filtering out non-biomedical claims from the dataset, and finally removing claims with incomplete and brief explanations.

Labels for news headline claims did not require standardization, as we assumed all news headline  claims (coming from reputable sources as they were) to be verified and thus labelled these \textit{true}, 
but filtered out from the dataset news entries with the headline prefixes ``AP EXCLUSIVE", ``Correction", ``AP Interview", and ``AP FACT CHECK". Indeed,  it would be difficult to  label the veracity of the claim in this type of entries. On the other hand, fact-check and news claims, which were associated with 141 different veracity labels, did require compression. We standardized the original labels for 4-way classification (see Appendix \ref{appendix:labels}). The chosen 4 labels are \textit{true}, \textit{false}, \textit{mixture}, and \textit{unproven}. We discounted claims with labels that cannot be reduced to one of these 4 labels. The  distribution of labels in the final \textsc{PubHealth} is shown in Table \ref{tab:data}. The dataset consists of a majority false claims. Unproven claims are the least common in the dataset.

\begin{table}[!htb]
\centering
\small
\begin{tabular}{lrrrrr}
\toprule

\textbf{Website}  & \textbf{tru.} & \textbf{fal.} &  \textbf{mix.} & \textbf{unp.} & \textbf{total}\\ 
\midrule
AP News &  2,132 & 0 & 0 & 0 & 2,132 \\ 

FactCheck & 0 & 50 & 29 & 8 & 87 \\ 

FullFact & 65 & 39 & 16 & 48 & 168 \\ 

HNR & 819 & 839 & 745 & 0 & 2,403  \\ 
  
Politifact & 671 & 1,339 & 423 & 0 & 2,433\\ 

Reuters & 1,971 & 0&  0 &  0 & 1,971 \\ 

Snopes & 386 & 1,131 & 405 & 220 &  2,142\\ 

TruthOrFict. & 132 & 172 & 120  & 72 & 496  \\ 
\midrule
Total & 6,176 & 3,570 & 1,526 & 299  &  11,832 \\
\bottomrule
\end{tabular}
\caption{Summary of the distribution of true (\textbf{tru.}), false (\textbf{fal.}), mixture (\textbf{mix.}) and unproven (\textbf{unp.}) veracity labels in \textsc{PubHealth}, across the original sources from which data originated.}
\label{tab:data}
\end{table}

The second step in processing the data was to remove claims with no biomedical context. This step was especially crucial for the claims which originated from fact-checking websites where the bulk of fact-checks concern political and economic claims. Health claims are easier to acquire from news websites, such as Reuters, as they can be quickly identified by the section of the website in which they were located during the data collection process. 
Although we mentioned that a sizeable number of claims from fact-checking sources are related to political events, some are connected to both political and health events or other mixed health context, and we collected claims whose subject matter intersects other topics in order to obtain a subject-rich dataset (see Appendix \ref{appendix:labels}).

Claims in the larger dataset were filtered according to a lexicon of 7,000 unique public health and health policy terms scraped from five health information websites (See Appendix \ref{appendix:labels}).

Furthermore, we manually added 65 more public health terms that were not retrieved during the initial scraping, but which we determined would positively contribute to the lexicon because of their relevance to the COVID-19 pandemic (see Appendix \ref{appendix:labels}). These claims were identified through exploratory data analysis of bigram and trigram collocations in \textsc{PubHealth}.

In order to filter out the entries which are not health-related, we kept only claims with main article texts that mentioned more than three unique terms in our lexicon.  Specifically, let $L$ be our lexicon, and $A_c$ and $T_c$, respectively, be the article text and claim text accompanying a candidate dataset entry $c$. Then, we included in \textsc{PubHealth} only the following set $C$ of claim entries, with accompanying information:

\begin{equation*}
\label{eq:1}
\begin{aligned}
C_{A} = \{c \mid 
\{l_1,...,l_n\}= A_{c}\cap L,
n > 3 \}
\end{aligned}
\end{equation*}

\begin{equation*}
\label{eq:2}
\begin{aligned}
C_{T} =\{c \mid 
\{l_1,...,l_n\}= T_{c}\cap L,
n > 3
\}
\end{aligned}
\end{equation*}

\begin{equation}
\label{eq:3}
\begin{aligned}
C = C_{A}\cup C_{T}
\end{aligned}
\end{equation}

As we already knew  that all Reuters health news claims qualify for our dataset, we used the lower bound frequency of words from our lexicon present in these article texts to determine our lower bound of three unique terms.
We acknowledge that there might be disparities in the amount of medical information present in entries. However, analysis of the dataset shows, quite promisingly, that on average  claims' accompanying article texts have $8.92\pm5.54$ unique health lexicon terms and claim texts carry $4.45\pm0.88$ unique terms 
from the health lexicon. 

Claims and explanations in the entries in the dataset were
also cleaned. Specifically, we also ensured all claims are between 25 and 400 characters in length. We removed explanations less than 25 characters long as we determined that very few claims shorter than this length contained fully formed claims; we removed claims longer than 400 characters to avoid the complexities of dealing with texts containing multiple claims. We also omitted claims and explanations ending in a question mark to ensure that all claims are statements, i.e., clearly defined.

Note that one aspect of the explanations' quality which we chose not to control, was the intended purpose of the text we labelled as the explanation: as shown in Table \ref{tab:explanation-sources} in Appendix~\ref{appendix:labels}, there was a wide variation across the websites we crawled. 

Table \ref{tab:comparisonfcdatasets} shows the Flesch-Kincaid~\cite{kincaid1975derivation} and Dale-Chall~\cite{chall1995readability} readability evaluations of claims from our \fc{} dataset when compared to four other fact-checking datasets.
The results show that \textsc{PubHealth} claims are, on average, the most challenging to read. Claims from our dataset have a mean Flesch-Kincaid reading ease score of 59.1, which corresponds to a 10th-12th grade reading level and fairly difficult to read. The other \fc{} datasets have reading levels which fit into the 6th, 7th and 8th grade categories.
Similarly for the Dale-Chall readability metric, on average our claims are more difficult to understand. Our claims have a mean score of 9.5 which is equivalent to the reading age of college student, whereas all other datasets' claims have an average score which indicates that they are readable by 10th to 12th grade students.
Both these results support our earlier assertion about the complexity of public health claims relative to political and more general claims.
 
\begin{table}[H]
    \centering
    \small
    \begin{tabular}{lccccl}
    \toprule
     \textbf{Dataset} & \multicolumn{2}{c}{\textbf{Flesch-Kincaid}} &  \multicolumn{2}{c}{\textbf{Dale-Chall}} \\
     \midrule
     & $\mu$  & $\sigma$  & $\mu$ & $\sigma$ \\
    \cmidrule(lr) {2-3}\cmidrule(lr) {4-5}
    
\footnotesize{\citeauthor{wang-2017-liar} (\citeyear{wang-2017-liar})} &  61.9 & 20.2 & 8.4 & 2.2 \\ 

\footnotesize{\citeauthor{shu2019fakenewsnet} (\citeyear{shu2019fakenewsnet})} & 67.1 & 24.3 & 8.9 & 3.0 \\

\footnotesize{\citeauthor{thorne-etal-2018-fever} (\citeyear{thorne-etal-2018-fever})} & 71.7 & 24.9 & 8.2 & 3.3 \\

\footnotesize{\citeauthor{augenstein-etal-2019-multifc} (\citeyear{augenstein-etal-2019-multifc})} & 60.8 & 22.1 & 8.9 & 2.5 \\
    \midrule
Our dataset &  59.1 & 23.3 & 9.5 & 2.6 \\

    \bottomrule
    \end{tabular}
    \caption{Comparison of readability of claims presented
in large fact-checking datasets (i.e., those with $> 10$K claims). We compute the mean and standard deviation for Flesch-Kincaid and Dale-Chall scores of claims for LIAR \cite{wang-2017-liar}, FEVER \cite{thorne-etal-2018-fever}, MultiFC \cite{augenstein-etal-2019-multifc}, FAKENEWSNET \cite{shu2019fakenewsnet}, and also our own \fc{} dataset. The sample sizes used for evaluation for each dataset are as follows, LIAR: 12,791, MultiFC: 34,842, FAKENEWSNET: 23,196, FEVER: 145,449, and 11,832 for our dataset.}
    \label{tab:comparisonfcdatasets}.
\end{table}

\section{Methods}\label{sec:methods}

In this section we describe in detail the methods we employed for devising automated fact-checking models.
We trained two \fc{} models: a classifier for veracity prediction, and a second summarization model for generating fact-checking explanations. The former returns the probability of an input claim text belonging to one of four classes: true, false, unproven, mixture. The latter uses a form of joint extractive and abstractive summarization to generate  explanations for the veracity of claims from article text about the claims. Full details of hyperparameters chosen and computer infrastructure which was employed can be found in Appendix \ref{appendix:reprodubility}.

\subsection{Veracity Prediction}

\begin{figure}[H]
    \centering
    \includegraphics[   width=0.48\textwidth,    
    trim = 0.2cm 0cm 0cm 0.2cm, clip]{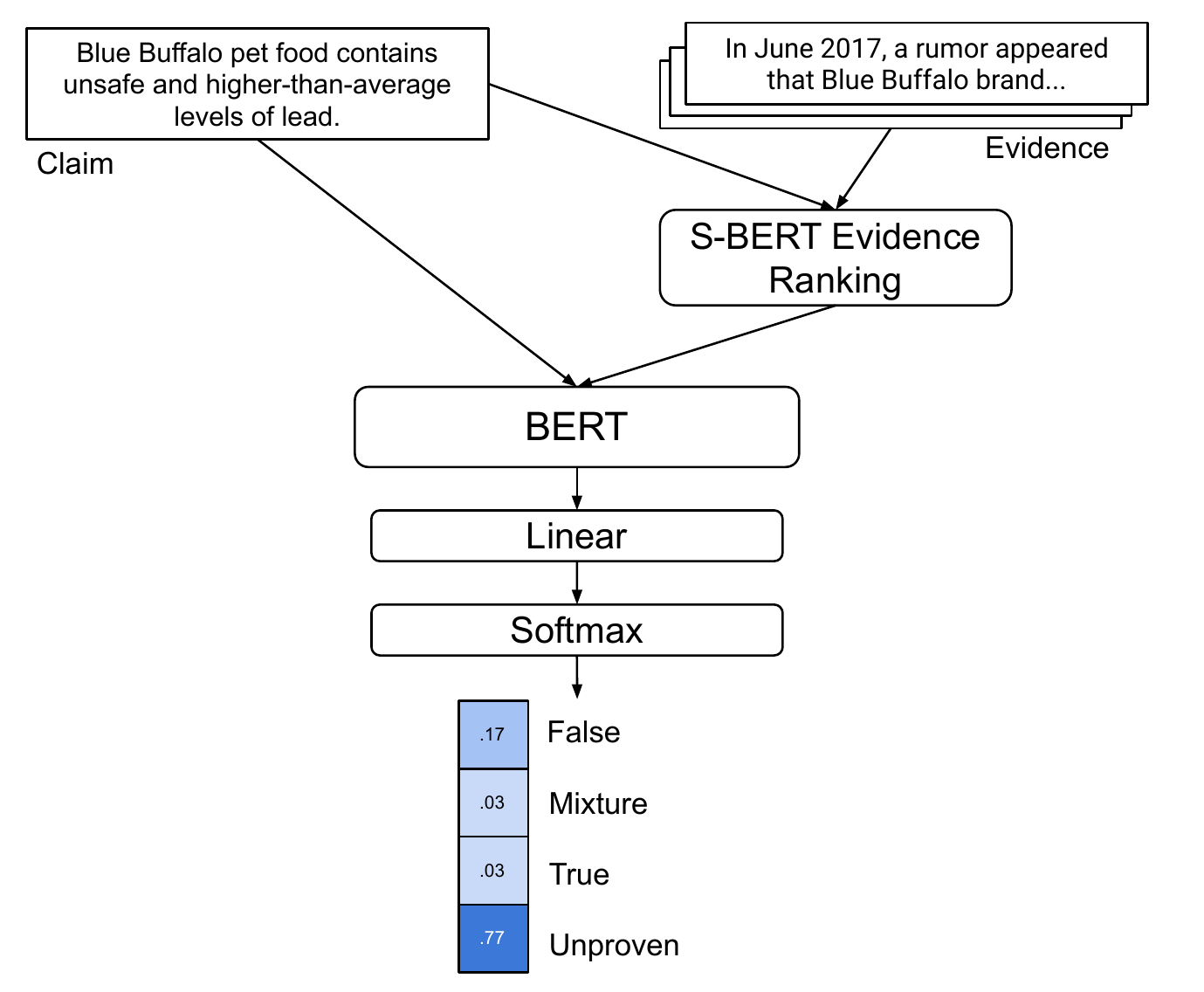}
    \caption{Architecture of veracity prediction.}
    \label{fig:pipeline}
\end{figure}

Veracity prediction is composed of two parts: evidence selection and label prediction (see Figure~\ref{fig:pipeline}).

For evidence selection, within fact-checking and news articles, we employ Sentence-BERT (S-BERT) \cite{reimers-gurevych-2019-sentence}. SBERT is a model for sentence-pair regression tasks which is based on the BERT language model \cite{devlin-etal-2019-bert}, to encode contextualized representations for each of the evidence sentences and then rank these sentences according to their cosine similarity with respect to the contextualized representation of the  claim sentence. We then 
select the top $k$ sentences for veracity prediction. As with sentence selection approaches from the fact-checking literature \cite{nie2019combining,zhong2019reasoning}, we choose $k=5$.

The claim and selected evidence sentences form the inputs for the label prediction part of our model (see Figure~\ref{fig:pipeline}). 
We fine-tuned, on the \textsc{PubHealth} dataset, pre-trained models for the downstream task of fact-checking label prediction. We employed four pre-trained models: original BERT uncased, \textsc{SciBERT}, \textsc{BioBERT} v1.0, and also \textsc{BioBERT} v1.1. 
The two versions of \textsc{BioBERT} differ slightly in that the earlier version is trained for 470K  steps on PubMed abstracts and PubMed Central (PMC) full article texts, whereas \textsc{BioBERT} v1.1 is trained for 1M steps on PubMed abstracts.

\subsection{Explanation Generation as Abstractive Summarization}

We make use of extractive-abstractive summarization \cite{Liu-Lapata:2019} in developing the explanation model. We choose this architecture because explanations for claims which concern a specific topic area having a highly complex lexicon   can benefit from the ability to articulate judgment in simpler terms. In order to deploy the model proposed by \cite{Liu-Lapata:2019}
we also implemented an explanation generation model.

Just as is the case for the predictor model, the explanation model is fine-tuned for the task on evidence sentences ranked by S-BERT. 
However, for the explanation model we use all article  sentences as well as the claim sentence to fine-tune a BERT-based summarization model pre-trained on the Dailymail/CNN news article and summaries dataset \cite{hermann2015teaching}. One of our models, \textsc{ExplainerFC}, is fine-tuned using non-public health data, which we extract from the portion of the 39.3K originally crawled fact-checks, news reviews, and news articles not included in \textsc{PubHealth}. For fairness, we ensure these data have the same proportion of claims from each website and the number of examples is the same as \textsc{PubHealth}. The second model, \textsc{ExplainerFC-Expert}, is fine-tuned  on \textsc{PubHealth}. Also, we evaluate both models on \textsc{PubHealth} test data.  Table \ref{tab:generated_explanations} shows an example of the explanations generated by the two methods. 

\section{Results}
We conducted experiments to evaluate the performance of both predictor(s) and explainer(s). The performance of the (various incarnations of the) prediction model is evaluated using an automatic approach, whereas the performance of the (two incarnations of the) explainer is assessed using both automatic and human evaluation.

\begin{figure}[ht]
\centering
\small
\begin{tikzpicture}
[
    auto,
    block/.style    = { rectangle,  
    draw=black, 
                        fill=white, text width=17em, text centered,
                        minimum height=2em },
    block1/.style    = { rectangle, draw=white,
                        fill=white, text width=17em, text centered,
                        minimum height=2em },
    line/.style     = { draw, thick, ->, shorten >=2pt },
  ]
 
  \matrix [column sep=5mm, row sep=0.5mm] {
                    
& \node [text centered] (x) {\textbf{Claim} \hspace{0.75cm} \textbf{Label:} \textcolor{alizarin}{\textbf{FALSE}}}; \\

& \node [block1] (doa) {Under Obamacare, patients 76 and older must be admitted to the hospital by their primary care physicians in order to be covered by Medicare.};  \\

& \node [text centered] (x) {}; \\
& \node [text centered] (x) {\textbf{Gold explanation}}; \\

& \node [block] (track) {Obamacare does not require that patients 76 and older must be admitted to the hospital by their primary care physicians in order to be covered by Medicare.}; & \\
& \node [text centered] (x) {}; \\
& \node [text centered] (y) {\textbf{EXPLAINERFC explanation}}; \\

& \node [block] (xfc)
                    {What's true: nothing in the Affordable Care Act requires that a primary care physician admit patients 76 or older to a hospital in order for their hospital care to be treated under Medicare. What's false: none of the provisions or rules put an upper age limit on medicare coverage.};            & \\ 
& \node [text centered] (x) {}; \\

& \node [text centered] (y) {\textbf{EXPLAINERFC-EXPERT explanation}}; \\

& \node [block] (xfc-expert)
                        {The Affordable Care Act does not require Medicare to admit patients to a hospital after paying the Part B deductible. It’s not the same age limit on medicare coverage. But the evidence doesn't specifically set an upper age limit.};          & \\
};

\end{tikzpicture}
\caption{Example of model-generated explanations as compared to the gold standard from our \fc{} dataset.
}

\label{tab:generated_explanations}
\end{figure}

\subsection{Prediction}

We split the \textsc{PubHealth} dataset as follows: 9,466 training examples, 1,183 examples for validation and 1,183 examples for testing. 

We evaluated veracity prediction using macro-F1, precision, recall and accuracy metrics as shown in Table \ref{tab:veracity_results}. We employ two baselines: a randomized sentence selection approach with BERT (bert-base-uncased) classifier, and lastly a BERT model, also using pre-trained uncased BERT, which does not make use of sentence selection and instead makes use of the entire article text to fine-tune for the fact-checking task. 

Out of the four BERT-derived models, \textsc{SciBERT} achieves the highest macro F1, precision and accuracy scores on the test set. \textsc{BioBERT} v1.1 achieves the second highest scores for F1, precision and accuracy. As expected, \textsc{BioBERT} v1.1 outperforms \textsc{BioBERT} v1.0 on all four metrics. The standard BERT model achieves the highest precision score of the four models, however it also achieves the lowest recall and F1 scores. This supports the argument we presented in Section \ref{sec:introduction} that subject-specific fact-checking can benefit from training on in-domain models.

\begin{table}[ht]
    \centering
    \small
    \begin{tabular}{lcccc}
    \toprule
   \textbf{Model}  & \textbf{Pr.} & \textbf{Rc.} & \textbf{F1} & \textbf{Acc.}\\
    \midrule
    BERT (rand. sents.) & 38.97 & 39.38 & 39.16 & 20.99\\
    BERT (all sents.) & 56.50 & 56.50 & 56.50 & 55.40\\
    \midrule
    \midrule
    \textsc{BERT} (top k sents.) & \textbf{77.39} & 54.77 & 63.93 & 66.02\\
    \textsc{SciBERT}  & 75.69 & \textbf{66.20} & \textbf{70.52} & \textbf{69.73}\\
    \textsc{BioBERT} 1.0 & 73.93 & 57.57 & 64.57 & 65.18\\ 
    \textsc{BioBERT} 1.1  & 75.04 & 61.68 & 67.48 & 68.89\\
    \bottomrule
    \end{tabular}
    \caption{Veracity prediction results for the two baselines and four BERT-based models on the test set. Model performance is assessed against precision (Pr.), recall (Rc.), macro F1, and accuracy (Acc.) metrics.}
    \label{tab:veracity_results}
\end{table}

\subsection{Explanations}

We use two methods for evaluating the quality of explanations generated by our methods: automated  evaluation and qualitative evaluation, in turn amounting to human and computational evaluation of explanation properties. 

\subsubsection{Automated Evaluation}
We make use of ROUGE summarization evaluation metrics  \cite{lin-2004-rouge}. Specifically we use the F1 values for ROUGE-1, ROUGE-2, and ROUGE-L, to evaluate  the explanations generated by the \textsc{ExplainerFC} and \textsc{ExplainerFC-Expert} models.

As in the setup employed by \citeauthor{Liu-Lapata:2019} (\citeyear{Liu-Lapata:2019}), we compare our explanation models to two other methods: a \textsc{Lead-3} baseline, which constructs a summary out of the first three sentences of an article, and an extractive summarization-based \textsc{Oracle} upper bound. 
The results of this evaluation are shown in Table \ref{tab:rouge}. The \textsc{ExplainerFC-Expert} explanation model outperforms \textsc{ExplainerFC}. \textsc{ExplainerFC-Expert} achieves higher scores than \textsc{ExplainerFC} for R1, R2, and RL metrics.

\begin{table}[ht]
    \centering
    \small
    \begin{tabular}{lccc}
    \toprule
    \multirow{2}{*}{Model}  & \multicolumn{3}{c}{ROUGE-F} 
    \\
       &  R1 & R2 & RL \\
     \midrule

     \textsc{Oracle}  & 39.24 & 14.89 & 32.78 \\
     \textsc{Lead-3} & 29.01 & 10.24 & 24.18\\
      \midrule
      \midrule
     \textsc{ExplainerFC} &  31.42 & 12.38 & 26.27 \\
     \textsc{ExplainerFC-Expert} & \textbf{32.30}  & \textbf{13.46} & \textbf{26.99}\\
    \bottomrule
    \end{tabular}
    \caption{ROUGE-1 (R1), ROUGE-2 (R2) and ROUGE-L (RL) F1 scores for explanations generated via our two explanation models.}
    \label{tab:rouge}
\end{table}

\subsection{Evaluation of Explanation Quality}
\label{sec:explanation-evaluation}
As the explanations we generate are from heterogeneous sources (and therefore not directly comparable), evaluation using ROUGE does not present us with a complete picture of the usefulness or quality of these explanations. For this reason, we adapt to the task of explainable fact-checking three of the desirable usability properties for machine learning explanations offered by \citeauthor{sokol2019desiderata} (\citeyear{sokol2019desiderata}). We define these properties  formally and evaluate the quality of the generated explanations against them. These same properties are also used for our human evaluations and for a comparison between human and computational evaluation of the quality of our explanations. To the best of our knowledge, ours is the first systematic evaluation of the quality of explanations for fact-checking in terms of formal properties. We define the three explanation properties as (two forms of) global coherence and (a form of) local coherence,
as follows.

\paragraph{Global Coherence} refers to the suitability of fact-checking explanations with respect to both the claim and label to which it is associated. We consider two incarnations of global coherence:
\begin{itemize}
    \item \emph{Strong global coherence}. Let $E$ be an explanation of the veracity label $l$ for claim $C$, where $
    e_{1},\dots, e_{N}
    $ are all the individual sentences which make up $E$. Then, $E$ satisfies strong global coherence iff $\forall e_{i}\in E$, $e_{i} \models C$. Put simply, for this property to hold for a generated fact-checking explanation, every sentence in the explanatory text must entail ($\models$) the claim. 
    
    \item \emph{Weak global coherence}. Let $E$ be an explanation of the veracity label $l$ for claim $C$, where $
    e_{1},\dots, e_{N}
    $ are all the individual sentences which make up $E$. Then, $E$ satisfies weak global coherence iff $\forall e_{i}\in E$,
    $e_{i} \not\models \neg C$. For this property to hold for a generated fact-checking explanation, no sentence in the explanatory text should contradict the claim (by entailing its negation); from a natural language inference (NLI) perspective, for  weak global coherence to hold all explanatory sentences should  entail or have a neutral relation with respect to the claim.
\end{itemize}

When measuring coherence, we treat as \textit{neutral}  claims originally labelled as  \textit{false} if their claim is \textit{contradicted} by its explanation. Note that if the \textit{false} claim is \emph{entailed} by its explanation  we do not reassign the label, because doing so would impose too strong an assumption that the entailment is related to the veracity which we cannot verify. 
  
\paragraph{Local Coherence.} Let $E$ be an explanation of the veracity label $l$ for claim $C$, where $
e_{1},\dots, e_{N}
$ are all the individual sentences which make up $E$. Then,  $E$ satisfies local coherence iff $\forall e_{i}, e_j \in E$,
$e_{i} \not\models \neg e_{j}$.

Local coherence is a measure of how cohesive sentences in an explanation are. For local coherence to hold any two sentences in an explanation must not contradict each other, i.e., there is no pairwise disagreement between sentences which make up the explanation.

Note that all three coherence properties relate to the usability property of coherence discussed by \citeauthor{sokol2019desiderata} (\citeyear{sokol2019desiderata}). Local coherence draws specifically on the idea of avoiding internal inconsistencies in explanations.
Figure~\ref{fig:text-example} shows an example of evaluation of the three properties, for a specific claim-explanation pair.
Schematic examples of explanations and evidence sentence relations which satisfy these coherence properties are shown in Appendix \ref{appendix:coherence}.

\subsubsection{Human \& Computational Evaluations}
We employ human evaluation in order to assess the quality of the gold and generated explanations with respect to these properties. Also, we conduct a computational evaluation of the three coherence properties using NLI.

For human evaluation, we randomly sampled 25 entries from the test set of \textsc{PubHealth}, and enlisted 5 annotators to evaluate the quality of the gold explanations and explanations generated by \textsc{ExplainerFC} and \textsc{ExplainerFC-Expert} for these entries.
We asked participants to annotate explanations according to the following criteria: 1) the agreement and disagreement between sentences in the explanation, and 2) relevance of the explanation to the claim. Further information, including an example from the questionnaire, can be found in Appendix \ref{appendix:questionnaire}.

We conducted the computational evaluation on three pretrained NLI models: 1) a decomposable attention model \cite{parikh-etal-2016-decomposable} using ELMo embeddings \cite{peters-etal-2018-deep} trained on 
the Stanford Natural Language Inference (SNLI) corpus \cite{Bowman:2015}, 2) RoBERTa \cite{liu2019roberta} trained on SNLI, and 3) RoBERTa trained on the Multi-Genre Natural Language Inference (MNLI) corpus \cite{williams-etal-2018-broad}. We implemented these evaluation methods using the AllenNLP platform \cite{gardner-etal-2018-allennlp}.  

For the human evaluation we computed Randolph's free-marginal $\kappa$ \cite{randolph2005free} and overall agreement (O.A.) for all multiple choice questions. For the gold explanations, we computed $\kappa$ (and O.A.) of 0.24 (62\%), 0.48 (65.6\%), and 0.39 (54.4\%) for 2-, 3-, and 4-ary questions respectively. For \textsc{ExplainerFC}, 0.06 (53.2\%), 0.17 (44.8\%), and 0.12 (34\%) for 2-, 3-, and 4-ary questions respectively. Lastly for \textsc{ExplainerFC-Expert}, we computed $\kappa$ and O.A. of 0.36 (68\%), 0.44 (62.73\%), and 0.20 (40\%) for 2-, 3-, and 4-ary questions. The computational evaluation was conducted on all examples from the test set. The results of both the  human and computational evaluation of the three coherence measures are shown in Table~\ref{tab:human}. Our results suggest that the NLI approximation is a reliable approximation for weak global coherence and local coherence properties. However, entailment appears to be a poor approximation for strong global coherence. Further, a larger human evaluation study would be required in order to verify these results.

\begin{table}[ht]
    \centering
 \small
\begin{tabular}{p{2.5cm}p{1cm}p{1cm}p{1cm}}
\toprule
\multirow{3}{*}{Evaluation Method} & \multirow{2}{*}{\textbf{SGC}} & \multirow{2}{*}{\textbf{WGC}} & 
\multirow{2}{*}{\textbf{LC}}\\

& & \\
 \cmidrule{2-4}
& \multicolumn{3}{c}{Gold explanations}\\
\midrule
Human & 76.80 & 98.40 & 65.60\\
DA+ELMo; SNLI & 8.72 & 87.61 & 55.20\\
RoBERTa; SNLI & 1.28 & 75.87 & 52.12\\
RoBERTa; MNLI & 2.66 & 87.52 & 54.84\\
\midrule
\multicolumn{4}{c}{\textsc{ExplainerFC} generated explanations}\\
\midrule
Human & 53.60 & 88.80  & 58.10\\
DA+ELMo; SNLI & 8.26 & 89.45 & 51.32\\
RoBERTa; SNLI & 0.46 & 76.42  & 48.01\\
RoBERTa; MNLI & 0.73 & 84.59 & 50.20\\

\midrule
\multicolumn{4}{c}{\textsc{ExplainerFC-Expert} generated explanations}\\
\midrule
Human & 60.4 & 76.80 & 59.30\\
DA+ELMo; SNLI & 7.61 & 89.72 & 64.60\\
RoBERTa; SNLI & 0.64 & 76.15 & 60.07\\
RoBERTa; MNLI & 2.48 & 84.04 & 62.43\\
       
\bottomrule
\end{tabular}
\caption{\% of explanations which satisfy strong global coherence (SGC), weak global coherence (WGC) and local coherence (LC) properties.}
\label{tab:human}
\end{table}

\begin{figure}[ht]
\begin{tabular}{|p{7.3cm}|}
\hline
        \textbf{Claim} \\
        A list of chemicals, written as if they were ingredients on a food label, accurately depicts the chemical composition of a banana. \\
        \textbf{Label: } \textcolor{darkgreen}{\textbf{TRUE}} \\
        \textbf{Explanation} \\
        In sum, this graphic accurately depicts the chemicals that comprise a banana, using a variety of tactics to make that completely natural food appear to be full of “chemicals” — something originally created by a high school chemistry teacher as part of a lesson on chemophobia. \\
        \hline
\end{tabular}
\caption{Example of 
explanation which satisfies all three coherence properties.}
\label{fig:text-example}
\end{figure}

\section{Conclusion and Future work}
In this paper, we explored fact-checking for claims for which specific expertise is required to produce a veracity prediction and explanations (i.e., judgments used for awarding the label/veracity prediction). To support this exploration we constructed \textsc{PubHealth}, a sizeable dataset for public health fact-checking and the first fact-checking dataset to include explanations as annotations. Our results show that  training veracity prediction and explanation generation models on in-domain data improves the accuracy of veracity prediction and the quality of generated explanations compared to training on generic language models without explanation. 

We hope to explore the topics of explainable fact-checking and specialist fact-checking further. In order to do this, we hope to explore other subjects, in addition to public health, for which fact-checking requires a level of expertise in the subject area. Furthermore, we hope to explore the quality of fact-checking explanations with respect to properties other than coherence, e.g., actionability and impartiality. lastly, we plan to explore congruity between veracity prediction and explanation generation tasks, i.e., generating explanations which are compatible with the predicted label and vice versa.

\section*{Acknowledgements}

We would like to thank all those who participated in the explanation evaluation study for their valuable contributions. 
The first author is supported by a doctoral training grant from the UK Engineering and Physical Sciences Research Council (EPSRC).

\bibliography{anthology,emnlp2020}
\bibliographystyle{acl_natbib}

\appendix
\section{Supplementary Material}




\subsection{Dataset}
\label{appendix:labels}
\begin{figure}[H]
\centering
\includegraphics[
width=0.49\textwidth,trim={0.0cm 0 0 0},clip]{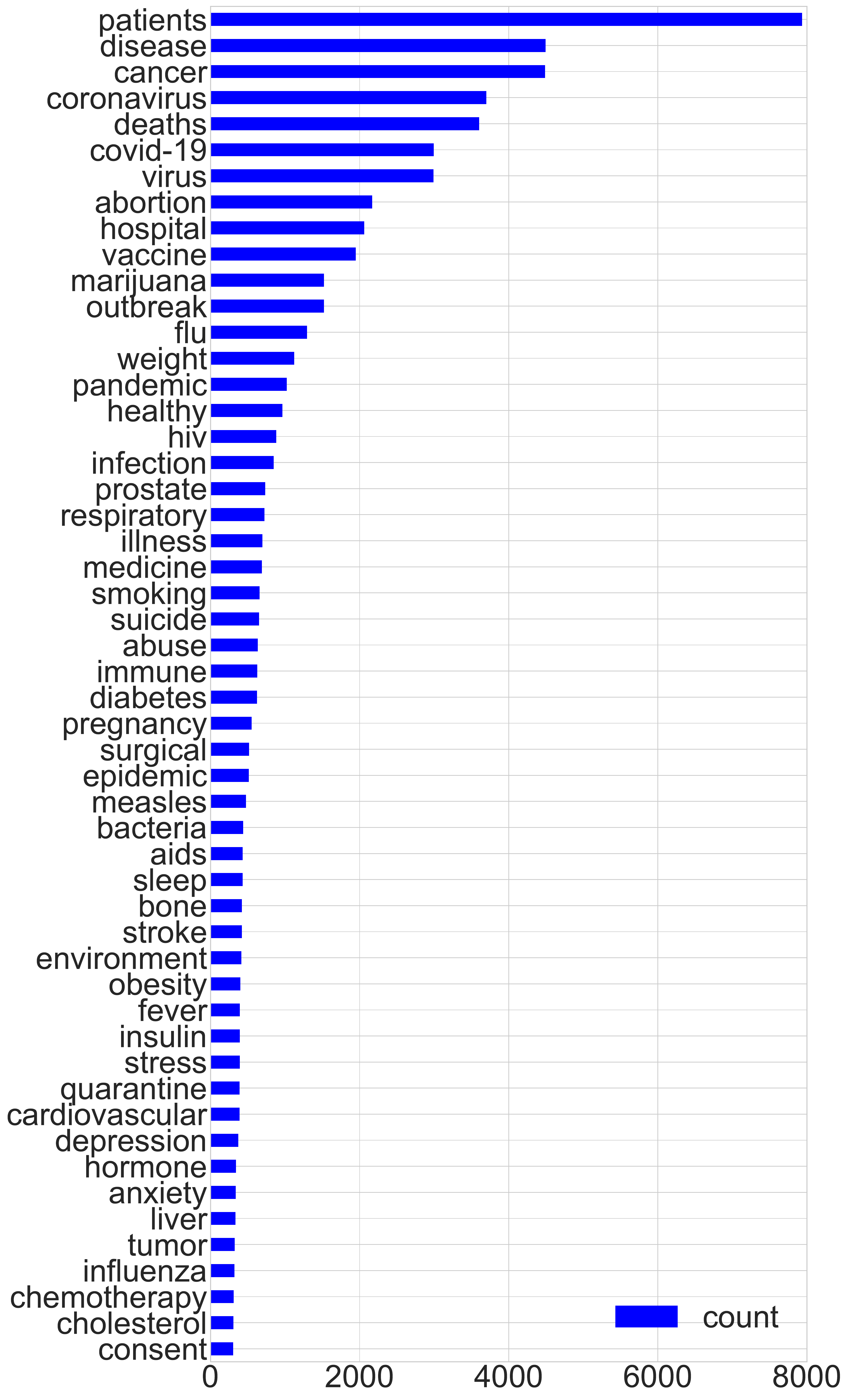}
\caption{Vocabulary from the health lexicon which features $>$ 300 times in \textsc{PubHealth} article texts.
}
\label{fig:freq_terms}
\end{figure}

\begin{figure}[H]
    \centering
    \includegraphics[
    width=0.5\textwidth,    
    trim = 0cm 0cm 0cm 2.5cm, clip]{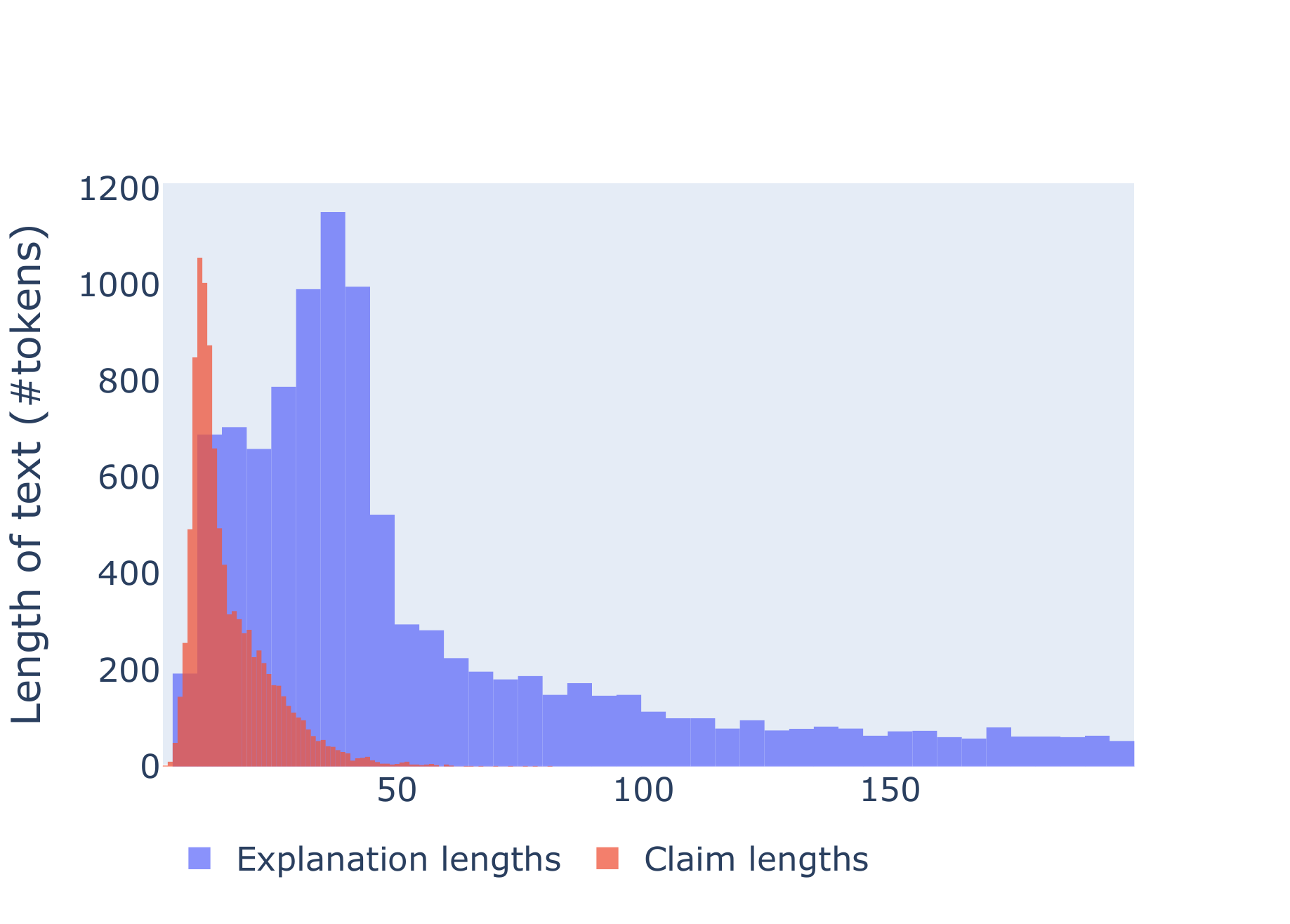}
    \caption{Histograms showing the distribution of lengths, measured by the number of tokens, for claims and explanations in the \textsc{PubHealth} dataset.}
    \label{fig:explanation_lengths}
\end{figure}

Here we expand on the dataset analysis presented in Section \ref{sec:data}. Figure \ref{fig:freq_terms} shows the most commonly occuring public health terms in the \textsc{PubHealth} dataset entry texts. Figure \ref{fig:explanation_lengths} illustrates the distribution of claim and explanation lengths. Note that the nature and format of the explanations for each of the scraped websites differed slightly.

Table \ref{tab:explanation-sources} shows the origin \fc{} explanations included in the \textsc{PubHealth} dataset. In Table \ref{tab:example-tags} we show examples of the subject-rich tags scraped alone with the claims. Table \ref{tab:labels} shows the  mapping between the standardized and original veracity labels.

\paragraph{Building the public health lexicon.} In order to compile the lexicon we scraped health related terms from the following website sources. In total we scraped vocabulary from a number of pages across six websites. These websites are NHS Health A-Z,\footnote{\url{https://www.nhs.uk/conditions/}} Everyday Health, \footnote{\url{https://www.everydayhealth.com/conditions/}} Medline Plus,\footnote{ \url{https://medlineplus.gov/encyclopedia.html}} Think Local, Act Personal,\footnote{\url{ https://www.thinklocalactpersonal.org.uk/Browse/Informationandadvice/CareandSupportJargonBuster/}}
National Careers Healthcare Job,\footnote{\url{ https://nationalcareers.service.gov.uk/job-categories/healthcare}} and 
the Mayo Clinic.\footnote{\url{ https://www.mayoclinic.org/diseases-conditions},  \url{https://www.mayoclinic.org/symptoms}, \url{https://www.mayoclinic.org/tests-procedures}, \url{https://www.mayoclinic.org/drugs-supplements}}

\paragraph{Additional words added the health lexicon.} The following are the extra words added to lexicon which we did not scraped. `Centers for Disease Control and Prevention', `abscess', `adolescence', `airborne', `alimentation', `alopecia', `aneurysm', `anorexia', `anti-vaxxer', `arrhythmia', `bacteria', `bacterium', `biohazard', `bioterrorism', `bleeding', `blood pressure', `chickenpox', `chloroquine', `contagious', `death', `disease', `embolism', `endemic', `environment', `epidemiology', `first aid', `flatten the curve', `flu', `gallbladder', `gangrene', `heart attack', `heparin', `hospital', `hydroxychloroquine', `hygiene', `hypertension', `illness', `immune', `infant mortality rate', `infect', `influenza', `lactose intolerance', `liver', `medicine', `menstruation', `mental health', `nurse', `organs', outbreak, pacemaker, `pandemic', `pathogen', `patients', `period poverty', `public health', `quarantine', `sickness', `smoking', `stroke', `surgical', `tumour', `vaccine', `ventilator', `virus', `x-ray'.

\begin{table}[H]
\centering
\small
\begin{tabular}{cp{4.5cm}}
\toprule
\textbf{Website}  & \textbf{Explanations}\\
\midrule
AP News (\textit{n}) & Leading paragraph. \\
FactCheck (\textit{f}) & Summarizing paragraph.\\
FullFact (\textit{f}) & Fact-check conclusions.  \\
HNR (\textit{r}) & Summary of reliability judment.  \\
Politifact (\textit{f}) & Fact-check ruling/rating comments. \\
Reuters (\textit{n}) & Leading paragraph.\\
Snopes (\textit{f}) & Fact-check what's true / false / undetermined or concluding paragraph.\\
TruthOrFict. (\textit{f}) & Summarizing paragraph.\\
\bottomrule
\end{tabular}
\caption{Format of explanations scraped from fact-checking (\textit{f}), news (\textit{n}), news review (\textit{r}) websites.}
\label{tab:explanation-sources}
\end{table}

\begin{table*}[ht]
    \centering
    \begin{tabular}{p{15cm}}
    \toprule
     \textbf{Claim:} Judge dismisses lawsuit over release of vaccination data.\\
    \textbf{Label: } \textcolor{darkgreen}{\textbf{TRUE}}\\
    \textbf{Tags:} Immunizations, Health, General News, Public health, Connecticut, Hartford, Bristol, Lawsuits\\
    \textbf{Date published}: September 30, 2019\\
    \hline 
    \textbf{Claim:} FDA allows marketing of cooling cap to reduce hair loss during chemotherapy.\\
    \textbf{Label: } \textcolor{brown}{\textbf{MIXTURE}}\\
    \textbf{Tags:} Breast cancer, FDA, medical devices, Women's health\\
    \textbf{Date published}: December 15, 2015\\
    \hline 
    \textbf{Claim:} Clinical study shows that retinal imaging may detect signs of Alzheimer's disease.\\\textbf{Label:} \textcolor{brown}{\textbf{MIXTURE}} \\
    \textbf{Tags:} Alzheimer's disease, NeuroVision Imaging LLC, retinal imaging\\
    \textbf{Date published}: 
    August 24, 2017\\
    \hline
    \textbf{Claim:} Salt lamps, because they emit negatively charged ions, impart myriad health benefits including reduced anxiety, improved sleep, increased energy, and protection from an “electric smog.”\\
    \textbf{Label}: \textcolor{alizarin}{\textbf{FALSE}} \\
    \textbf{Tags:} medical, salt lamps \\ 
    \textbf{Date published}:  December 22, 2016\\
    \bottomrule
    \end{tabular}
    \caption{Examples of tag metadata for entries in the \textsc{PubHealth} dataset.}
    \label{tab:example-tags}
\end{table*}   

\begin{table*}[ht]
\centering
\begin{tabular}{cp{12cm}}
\toprule
\textbf{Standardized}  & \textbf{Fact-checking and news review veracity labels}\\
\midrule
\textbf{false} &  `0 Star', `1 Star', `2 Star', `barely-true', `digital manipulations!', `disputed', `disputed!', `false', `fiction', `fiction!', `fiction! \& disputed!', `fiction! \\& satire!', `full-flop', `inaccurate attribution!', `incorrect attribution!', `incorrect authorship!', `incorrectly attributed!', `misattributed', `mostly fiction!', `mostly-false', `not true', `pants-fire', `pants-on-fire!', `reported as fiction!', `reported fiction! \\[0.2cm] 
\textbf{mixture} &  `3 Star', `cherry picks', `confirmed authorship! but inaccurate attribution!', `decontextualized', `depends on where you vote!', `distorts the facts', `exaggerates', `half-flip', `half-true', `lacks context', `misleading', `misleading!', `mixed', `mixture', `not the whole story', `outdated', `outdated!', `previously truth! \& now resolved!', `previously truth! but now resolved!', `reported as truth! \& disputed!', `spins the facts', `truth \& fiction!', `truth! \& disputed!', `truth! \& fiction!', `truth! \& fiction! \& disputed!', `truth! \& fiction! \& unproven!', `truth! \& misleading!', `truth! \& outdated!', `truth! \& unproven!', `truth! and fiction!', `truth! and unproven!', `truth! but decision reversed!', `truth! but inaccurate description!', `truth! but misleading!', `truth! but obama quote is fiction!', `truth! but overturned!', `truth! but resolved!', `truth! but she denies it reflects her views!', `truth! fiction! \& disputed!', 'truth! fiction! \& satire!', `truth! fiction! \& unproven!', `truth!, fiction!, and unproven!', `truth!, unproven!, \& fiction!'
\\[0.2cm] 
\textbf{true} & `4 Star', `5 Star', `authorship confirmed!', `commentary!', `confirmed authorship', `confirmed authorship!', `correct attribution!', `correct-attribution', `correctly attributed!', `mostly truth!', `mostly-true', `no-flip', `official!', `reported to be true!', `reported to be truth!', `true', `truth but an opinion!', `truth!', `truth! but an opinion!', `truth! but not intentionally!', `truth! but not the one you think!', `truth! but now resolved!' \\[0.2cm] 
\begin{tabular}{c} \textbf{unproven} \end{tabular}& `investigation pending!', `no evidence', `pending investigation!', `unconfirmed attribution!', `unknown', `unofficial!', `unproven', `unproven!', `unsupported' \\
\bottomrule
\end{tabular}
\caption{These are the four standardized labels we defined for veracity prediction (left) and lists (right) of the original fact-checking labels provided by the fact-checking and news review websites we scraped, mapped to our four standardized labels}.
\label{tab:labels}
\end{table*}

\subsection{Reproducibility}
\label{appendix:reprodubility}
Here we provide further information about the experiments described in Section \ref{sec:methods}.

\paragraph{Prediction models hyperparameters.} We perform hyper-parameter grid search as part of validation for batch sizes from \{8, 16, 32\}, learning rates from \{1e-5, 5e-6, 1e-6\}, and epochs \{2, 3, 4\}. We optimize our veracity prediction model on cross entropy loss. The hyper-parameters we selected from this grid search are a batch size of 16, learning rate 1e-6 and 4 epochs for model training.

\paragraph{Computing Infrastructure.} All experiments were run on a machine with a dual Intel(R) Core(TM) i9-9900X 3.50GHz CPU. The GPU used for experiments is the Nvidia GeForce RTX 2080 Ti model. Additional information about the software packages used in the development of the explanation generation and veracity prediction models can be found in the GitHub repository, the link to which is given in Footnote 1.

\subsection{Human Evaluation Questionnaire}
\label{appendix:questionnaire}

The following are example question and response pairs typical of those presented to participants in the human evaluation questionnaire (see Section \ref{sec:explanation-evaluation}). Question and response pairs are related to the claim and explanation presented below.

\begin{enumerate}
    \item \textbf{Question}: Are there any sentences or phrases in the  explanation which disagree with each other?

\textbf{Response options}: \{Yes, No\}.

\item \textbf{Question}: Which veracity label would you give to the claim taking into account the entire explanation?

\textbf{Response options}: \{Mixture, false, true, unproven\}.
\end{enumerate}


\begin{tabular}{|p{6.7cm}|}
        \hline
        \textbf{Claim} \\
        State reports new findings of mosquito-borne illnesses. \\
        \textbf{Explanation} \\
        Rhode Island health officials say a second mosquito case tested positive for eastern equine encephalitis has been confirmed in the state, marking the first human case of the equine encephalitis in Rhode Island in more than two years. \\
        \hline
\end{tabular}

\subsection{Coherence properties}
\label{appendix:coherence}

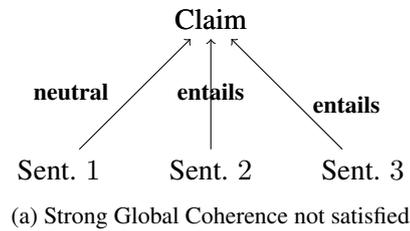
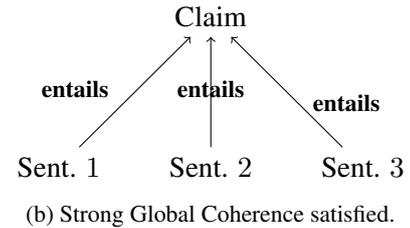
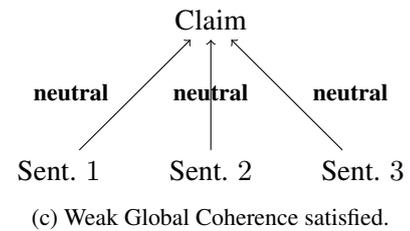
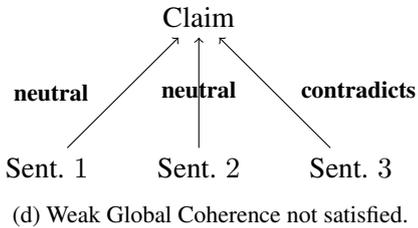
\begin{figure}[htb!]
\centering
\begin{subfigure}[b]{0.5\textwidth}
\centering
\begin{tikzpicture}
\node[label] (ArgumentA) at (5,1) {Claim};
\node[label] (ArgumentA) at (5,1) {Claim};
  \node[text=black] (ArgumentC) at (3,-1) {Sent. $1$};  
  \node[text=black] (ArgumentD) at (5,-1) {Sent. $2$}; 
\node[text=black] (ArgumentB) at (7,-1) {Sent. $3$};
  \draw[->,draw=black] (ArgumentB) -- (ArgumentA) node[midway, right] {\small \begin{tabular}{c}\\\textbf{entails}\end{tabular}};
  \draw[->,draw=black] (ArgumentC) -- (ArgumentA) node[midway, left] {\small
  \begin{tabular}{c}\textbf{neutral}\end{tabular}};
\draw[->,draw=black] (ArgumentD) -- (ArgumentA)
node[midway] {\small
  \begin{tabular}{c}\textbf{entails}\end{tabular}};
\end{tikzpicture}
\caption{Strong Global Coherence not satisfied}
\label{fig:sgc-notsatisfied}
\end{subfigure}

\begin{subfigure}[b]{0.5\textwidth}
\centering
\begin{tikzpicture}
\node[label] (ArgumentA) at (5,1) {Claim};
  \node[text=black] (ArgumentC) at (3,-1) {Sent. $1$};  
  \node[text=black] (ArgumentD) at (5,-1) {Sent. $2$}; 
\node[text=black] (ArgumentB) at (7,-1) {Sent. $3$};
  \draw[->,draw=black] (ArgumentB) -- (ArgumentA) node[midway, right] {\small \begin{tabular}{c}\\\textbf{entails}\end{tabular}};
  \draw[->,draw=black] (ArgumentC) -- (ArgumentA) node[midway, left] {\small
  \begin{tabular}{c}\textbf{entails}\end{tabular}};
\draw[->,draw=black] (ArgumentD) -- (ArgumentA)
node[midway] {\small
  \begin{tabular}{c}\textbf{entails}\end{tabular}};
\end{tikzpicture}
\caption{Strong Global Coherence satisfied.}
\label{fig:sgc-satisfied}
\end{subfigure}

\begin{subfigure}[b]{0.5\textwidth}
\centering
\begin{tikzpicture}
\node[label] (ArgumentA) at (5,1) {Claim};
  \node[text=black] (ArgumentC) at (3,-1) {Sent. $1$};  
  \node[text=black] (ArgumentD) at (5,-1) {Sent. $2$}; 
\node[text=black] (ArgumentB) at (7,-1) {Sent. $3$};
  \draw[->,draw=black] (ArgumentB) -- (ArgumentA) node[midway, right] {\small \begin{tabular}{c}\textbf{neutral}\end{tabular}};
  \draw[->,draw=black] (ArgumentC) -- (ArgumentA) node[midway, left] {\small
  \begin{tabular}{c}\textbf{neutral}\end{tabular}};
\draw[->,draw=black] (ArgumentD) -- (ArgumentA)
node[midway] {\small
  \begin{tabular}{c}\textbf{neutral}\end{tabular}};
\end{tikzpicture}
\caption{Weak Global Coherence  satisfied.}
\label{fig:wgc-notsatisfied}
\end{subfigure}

\begin{subfigure}[b]{0.5\textwidth}
\centering
\begin{tikzpicture}
\node[label] (ArgumentA) at (5,1) {Claim};
  \node[text=black] (ArgumentC) at (3,-1) {Sent. $1$};  
  \node[text=black] (ArgumentD) at (5,-1) {Sent. $2$}; 
\node[text=black] (ArgumentB) at (7,-1) {Sent. $3$};
  
  \draw[->,draw=black] (ArgumentB) -- (ArgumentA) node[midway, right] {\small \begin{tabular}{c}\textbf{{contradicts}}\end{tabular}};
  \draw[->,draw=black] (ArgumentC) -- (ArgumentA) node[midway, left] {\small
  \begin{tabular}{c}\colorbox{white}{\textbf{neutral}}\end{tabular}};
\draw[->,draw=black] (ArgumentD) -- (ArgumentA)
node[midway] {\small
  \begin{tabular}{c}\textbf{neutral}\end{tabular}};
\end{tikzpicture}
\caption{Weak Global Coherence not satisfied.}
\label{fig:wgc-satisfied}
\end{subfigure}

\caption{Schematic representations of strong and weak global coherence properties.
\label{fig:coherence properties}}
\end{figure}

Figure \ref{fig:coherence properties} shows  examples of the three coherence properties mentioned in Section \ref{sec:explanation-evaluation}, shown schematically in graphical form.

\begin{figure}[ht]
    \centering
\begin{subfigure}[b]{0.5\textwidth}
\centering
\begin{tikzpicture}
\node[label] (ArgumentA) at (5,1) {Sent. $1$};
  \node[text=black] (ArgumentC) at (3,-1) {Sent. $2$};  
\node[text=black] (ArgumentB) at (7,-1) {Sent. $3$};
  \draw[<->,draw=black] (ArgumentB) -- (ArgumentA) node[midway, right] {\small \begin{tabular}{c}\textbf{neutral}\end{tabular}};
  \draw[<->,draw=black] (ArgumentC) -- (ArgumentA) node[midway, left] {\small
  \begin{tabular}{c}\textbf{neutral}\end{tabular}};
\draw[<->,draw=black] (ArgumentB) -- (ArgumentC) node[midway, above]
{\small
  \begin{tabular}{c}\textbf{neutral}\end{tabular}};
\end{tikzpicture}
\caption{Local coherence satisfied.}
\label{fig:local-satisfied}
\end{subfigure}

\begin{subfigure}[b]{0.5\textwidth}
\centering
\begin{tikzpicture}
\node[label] (ArgumentA) at (5,1) {Sent. $1$};
  \node[text=black] (ArgumentC) at (3,-1) {Sent. $2$};  
\node[text=black] (ArgumentB) at (7,-1) {Sent. $3$};
  \draw[<->,draw=black] (ArgumentB) -- (ArgumentA) node[midway, right] {\small \begin{tabular}{c}\textbf{neutral}\end{tabular}};
  \draw[<->,draw=black] (ArgumentC) -- (ArgumentA) node[midway, left] {\small \begin{tabular}{c}\textbf{neutral}\end{tabular}};
\draw[<->,draw=black] (ArgumentB) -- (ArgumentC) node[midway, above]
{\small \begin{tabular}{c}\textbf{contradicts}\end{tabular}};
\end{tikzpicture}
\caption{Local coherence not satisfied.}
\label{fig:local-nsatisfied}
\end{subfigure}
    \caption{Schematic representations of local coherence.}
    \label{fig:coherence properties 2}
\end{figure}
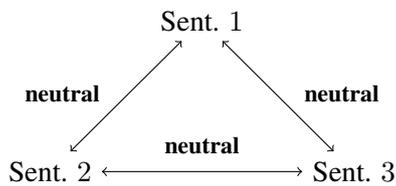
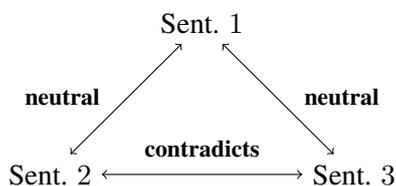

\end{document}